\documentclass[10pt,twocolumn,letterpaper]{article}

\usepackage{cvpr}
\usepackage{times}
\usepackage{epsfig}
\usepackage{graphicx}
\usepackage{amsmath}
\usepackage{amssymb}
\usepackage{color}
\usepackage[labelsep=period]{caption}
\usepackage{subfigure}
\usepackage{caption}
\usepackage{cite}

\usepackage{url}
\usepackage[toc,page]{appendix}
\usepackage{pifont}
\usepackage[pagebackref=true,breaklinks=true,letterpaper=true,colorlinks,bookmarks=false]{hyperref}
\usepackage{array}
\usepackage{enumitem}

\newcommand{\Cgray}[1]{\cellcolor[HTML]{D8D6D6}} %

\def\cvprPaperID{1633} 

\cvprfinalcopy

\begin{document}
	
	\title{Distance-based Camera Network Topology Inference for Person Re-identification}
	
	\author{\qquad Yeong-Jun Cho and Kuk-Jin Yoon\\
		\qquad Computer Vision Laboratory, GIST, South Korea\\
		{\tt\small\qquad $\lbrace${yjcho, kjyoon}$\rbrace$@gist.ac.kr}
		\and
	}
	
	\maketitle
	
	\begin{abstract}
		
		In this paper, we propose a novel distance-based camera network topology inference method for efficient person re-identification. To this end, we first calibrate each camera and estimate relative scales between cameras. Using the calibration results of multiple cameras, we calculate the speed of each person and infer the distance between cameras to generate distance-based camera network topology. The proposed distance-based topology can be applied adaptively to each person according to its speed and handle diverse transition time of people between non-overlapping cameras. To validate the proposed method, we tested the proposed method using an open person re-identification dataset and compared to state-of-the-art methods.
		The experimental results show that the proposed method is effective for person re-identification in the large-scale camera network with various people transition time.
		
	\end{abstract}
	
	
	\vspace{-10pt}
	\section{Introduction}
	
	Numerous surveillance cameras installed in public places (e.g., offices, stations, and streets) allow monitoring and tracking of people in large-scale environments.
	However, it is difficult for a person to individually observe each camera. To reduce the human efforts, person re-identification techniques which automatically identify people between multiple non-overlapping cameras can be used.
	Previous works have mainly focused on modeling people appearance information such as feature descriptor extraction~\cite{farenzena2010person, liu2012person} and learning similarity metrics~\cite{koestinger2012large,dikmen2011pedestrian} to perform person re-identification.
	Recently, many appearance modeling methods based on deep neural network~\cite{ahmed2015improved, yi2014deep} have been proposed.
	
	However, in a large-scale camera network where the number of cameras and the number of people are large, the person re-identification problem becomes very challenging. 
	In particular, it is difficult to effectively perform re-identification only by using the appearance-based methods in the large-scale camera network.
	This is because the conventional appearance-based methods do not take the structure of the camera network (e.g., spatial and temporal connectivities between cameras) into account; therefore, they have to examine every possible person candidate in the camera network to re-identify a person. 
	Instead of examining every person, we can restrict and reduce the search space by inferring the spatio-temporal relation between cameras, referred to as a camera network topology.
	
	\begin{figure}[t]	
		\centering
		\includegraphics[width=0.95\columnwidth]{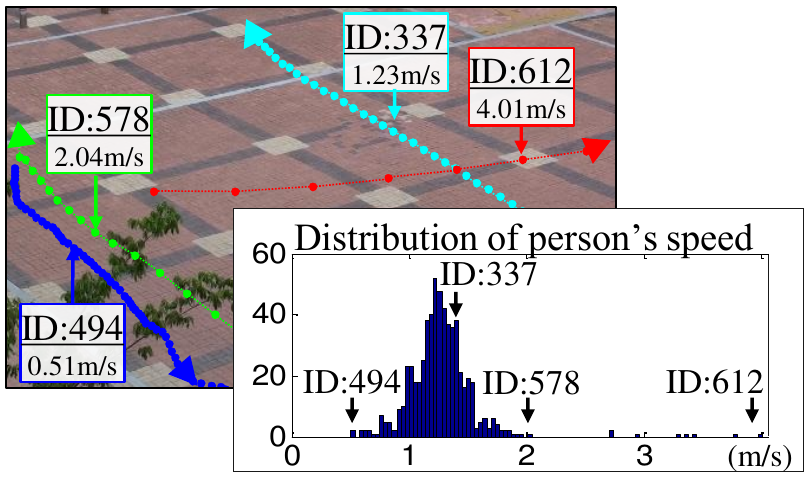}
		\caption{Challenges in person re-identification based on the time-based camera network topology due to the diverse speeds of people. Each blob was marked at every 0.3 seconds interval and each color indicates a person identity.}
		\label{fig_1}
		\vspace{-10pt}
	\end{figure}
	
	In recent years, several camera network topology inference methods~\cite{makris2004bridging, stauffer2005learning, chen2014object} have been proposed.
	In general, camera network topology represents spatio-temporal relations and connections between cameras. 
	The topology is represented as a graph $G_t = (V,E_t)$, where vertices $V$ denote cameras, and edges $E_t$ denote the transition distribution of people across cameras according to `time'. We name this topology $G_{t}$ as a time-based camera network topology.
	Although there has been much progress in person re-identification using camera network topology, the time-based topology has difficulty in dealing with the diverse walking speed of the people.
	As shown in Fig.~\ref{fig_1}, the walking speed of people is very diverse. For example, if a person walks much faster or slower than the average speed, it is hard to predict its transition time between two non-overlapping cameras based on the time-based topology. 	
	To overcome the limitations of the previous works, we propose a novel \emph{distance}-based camera network topology inference method to perform efficient and accurate person re-identification.
	
	To estimate the speed of a moving person, we need to calibrate cameras in a network automatically. 
	The progress in self-calibration methods~\cite{lv2006camera, liu2011surveillance} enables us to estimate camera parameters without any off-line calibration steps~\cite{zhang1999flexible}.	However, the estimated camera extrinsic parameters (i.e., camera position $\mathbf{t}$) are not accurate due to scale ambiguity in the self-calibration method, \ie we can obtain the camera extrinsic parameters determined up to scale. 
	In this work, we first estimate the relative scale between cameras based on human height information and correct the inaccurate camera calibration results.
	We then calculate speeds of all people in the camera network.  
	Subsequently, we infer the distance between cameras by multiplying the speeds and transition times of people and build a distance-based camera network topology.
	The inferred distance-based topology can be applied adaptively to each person -- when we divide the distance-based camera network topology according to the speed of a person, it gives a person-specific time-based camera network topology.
	
	The main idea of this work is simple but effective. To the best of our knowledge, this is the first attempt to infer the distance-based camera network for person re-identification.
	To validate the proposed method, we tested the proposed method using the \texttt{SLP}~\cite{Cho_2017_ICCV_Workshops} dataset and compared with state-of-the-art methods. The results show that the proposed method is promising for person re-identification in the large-scale camera network with diverse speeds of people.
	
	The rest of the paper is organized as follows: In Sec.~\ref{sec:preivous}, we review previous works of person re-identification and camera network topology inference. In Sec.~\ref{sec:proposed}, we describe our proposed distance-based camera network topology inference and person re-identification methods. The dataset and evaluation methodologies are described in Sec~\ref{sec:data_meth}. The experimental results are reported in Sec.~\ref{sec:exp} and we conclude the paper in Sec.~\ref{sec:conclusion}.

	\section{Previous Works}
	\label{sec:preivous}
	

	\subsection{Person Re-identification}
	In general, most of the person re-identification methods rely on appearances of people to identify people across non-overlapping views.
	To describe and classify the appearances of people, many works have tried to propose appearance modeling methods such as feature learning and metric learning methods.
	For the feature learning, \cite{farenzena2010person, liu2012person,zhao2014learning} designed feature descriptors to well describe the appearance of people.
	For the metric learning, many methods~\cite{koestinger2012large,dikmen2011pedestrian} have been proposed and used for person re-identification~\cite{davis2007information, weinberger2005distance, roth2014mahalanobis}. Several works~\cite{koestinger2012large,roth2014mahalanobis} extensively evaluated the feature and metric learning methods to show the effectiveness of those methods in the person re-identification task.
	In addition, several works exploited additional information such as human pose prior~\cite{wu2015viewpoint, cho2016improving, Su_2017_ICCV} and group appearance model~\cite{Zheng2009associating, Lisanti_2017_ICCV} to improve the re-identification performance.
	
	In recent years, many re-identification methods based on learning deep convolutional neural network (CNN)~\cite{zheng2016mars, su2016deep, xiao2016learning} and Siamese convolutional network~\cite{ahmed2015improved, yi2014deep, wang2016joint, Chung_2017_ICCV} have been proposed for simultaneously learning both features and metrics.
	In addition, several works utilized recurrent neural network (RNN)~\cite{mclaughlin2016recurrent, Zhou_2017_CVPR_P2S, Zhou_2017_CVPR_forest} and long short term memory (LSTM)~\cite{yan2016person} network to perform multi-shot person re-identification.
	Although a lot of person re-identification methods have been proposed so far, the challenges of person re-identification in a large-scale camera network, e.g., spatio-temporal uncertainty between non-overlapping cameras and high computational complexity, still remain.

	\begin{figure*}[t]	
		\centering
		\includegraphics[width=2.1\columnwidth]{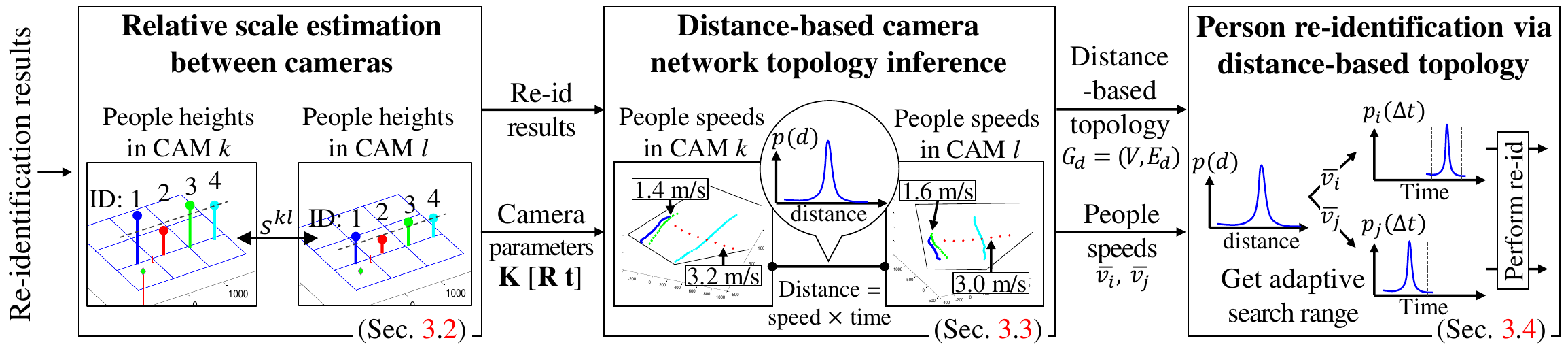}
		\caption{Overview of the proposed framework: distance-based camera network topology inference and person re-identification.}
		\label{fig_2} \vspace{-5pt}
	\end{figure*}
	
	\subsection{Camera Network Topology Inference} 
	
	Recently, many works have tried to infer a camera network topology and employ camera geometry to resolve the spatio-temporal ambiguities. 	
	Several works~\cite{javed2003tracking,cai2014exploring,rahimi2004simultaneous} assumed the camera network topology is given and showed the effectiveness of the spaito-temporal information between cameras.
	However, in practice, the camera network topology is not given. Thus, many works have tried to infer the camera network topology in an unsupervised manner.
	Makris~\textit{et al.}~\cite{makris2004bridging} proposed a topology inference method based on a simple event correlation model between cameras. 
	This topology inference method was extended in many works~\cite{niu2006recovering,stauffer2005learning,chen2014object}.
	Similarly, Loy~\textit{et al.}~\cite{loy2010time,loy2012incremental} inferred a camera network topology by measuring mutual information between activity patterns of  cameras.
	
	The previous topology inference methods~\cite{makris2004bridging, stauffer2005learning,chen2014object, loy2010time,loy2012incremental} are practical since they do not require appearance matching steps such as re-identification or inter-camera tracking for topology inference.
	However, the inferred topologies are prone to be inaccurate since the topology is easily contaminated by false event correlations, which frequently occur when people pass through blind regions irregularly.
	On the other hand, several works~\cite{cai2010recovering,martinel2016person, Cho_2017_ICCV_Workshops} inferred the camera network topology using person re-identification results. These methods can be more robust to noise than the event-based approaches since the methods infer the topology by utilizing true correspondences between cameras.
	Especially, Cho~\textit{et al.}~\cite{Cho_2017_ICCV_Workshops} iteratively solved re-identification and camera network topology inference. It achieved accurate results in both tasks thanks to its iterating strategy.
	
	The previous camera network topology inference methods, which have been proposed so far, mainly focused on to infer the transition time of people between cameras.
	However, the previous methods, so-called, `time-based topology' inference methods cannot efficiently handle the diverse speeds of people as shown in Fig.~\ref{fig_1}.

	\section{Proposed Methods}
	\label{sec:proposed}
	
	As mentioned above, many camera network topology inference methods have been proposed recently to perform efficient person re-identification in a large-scale camera network.
	They inferred a time-based camera network topology based on the people transition time between cameras. However, the speed of people can be very diverse, and it leads diverse people transition time between cameras as shown in Fig.~\ref{fig_1}.
	Therefore, the conventional time-based topology becomes ambiguous and inaccurate under the diverse speed of people.
	
	Actually, in surveillance videos, it is possible to extract additional cues such as camera parameters (position and viewpoint) and trajectories of people.
	Using the additional cues, we can also estimate walking speed of each person. Then, the challenge of re-identification due to diverse walking speed becomes more tractable.
	In this work, we fully exploit those additional cues and propose a new distance-based camera network topology inference, which does not depend on the speed of a person.
	
	\subsection{Overall Proposed Framework}
	\label{subsec:overall}
	
	We first obtain initial person re-identification results (i.e., person correspondence pairs) between cameras.
	To this end, we can apply any existing re-identification methods except for methods using prior knowledge of the camera network\footnote{For example, person re-identification based on metric learning~\cite{koestinger2012large, dikmen2011pedestrian} requires true person correspondences between cameras to learn the distance metric. We aim to run our framework without any prior knowledge of the camera network. Thus we do not use the methods requiring prior knowledge of the camera network.}.
	The initial re-identification results are utilized in following proposed steps.
	In Sec.~\ref{subsec:scale}, we perform the relative scale estimation for each camera. Each camera in the camera network is calibrated using camera self-calibration techniques~\cite{liu2011surveillance, lv2006camera} and its camera parameters are adjusted based on the proposed method.
	After estimating relative scales between cameras, we calculate walking speeds of people in each camera. The speeds of people and re-identification results are used to infer the distance-based camera network topology as described in Sec~\ref{subsec:dist_top}.
	Finally, we perform person re-identification based on the inferred distance-based camera network topology in Sec.~\ref{subsec:re_id}.
	The proposed framework is illustrated in Fig.~\ref{fig_2}.
	For reproducibility, the code of this work is available to the public at: {\url{https://}}.

	\subsection{Relative Scale Estimation between Cameras}
	\label{subsec:scale}
	
	In a general pinhole camera model, the relation between a 2D image (pixel coordinates $[u,v]$) and a 3D point (world coordinates $[X,Y,Z]$) can be represented by 3$\times$ 4 projection matrix $\mathbf{P}$ as 
	\begin{equation}
	\begin{split}
	\left[ u, v, 1 \right]^{\top} & = \mathbf{P} \left[ X, Y, Z, 1 \right]^{\top}, \\
	\mathbf{P} &= \mathbf{K}\left[ \mathbf{R} \quad \mathbf{t} \right],
	\end{split}
	\end{equation}
	where $\mathbf{K}$, $\left[\mathbf{R} \quad \mathbf{t}\right]$ represent camera intrinsic and extrinsic parameters.
	Unfortunately, most surveillance cameras remain uncalibrated.
	
	In order to estimate the camera parameters, camera self-calibration techniques~\cite{liu2011surveillance, lv2006camera} can be employed. These methods do not require any pre-defined checkerboards and off-line calibration tasks~\cite{zhang1999flexible}. 
	Instead of using a checkerboard, they utilize a human height $H$ as the checkerboard.
	A camera extrinsic parameter $\mathbf{t}$ (camera position) is determine by the value of $H$.
	However, $H$ is unknown in general; thus Liu \textit{et al.}~\cite{liu2011surveillance} set $H$ to any pre-defined specific value (e.g., the average height of humans: $H$=1.72m~\cite{visscher2008sizing}). 
	Therefore, although we can estimate the intrinsic and extrinsic parameters of each camera using the self-calibration technique, the camera extrinsic parameter $\mathbf{t}$ is not accurate since we use the inaccurate $H$ value. 
	In this work, we consider every camera in the camera network; thus each camera extrinsic parameter $\mathbf{t}$ should be adjusted to share the same world coordinate system.
	
\begin{figure}[t]	
		\centering
		\includegraphics[width=0.99\columnwidth]{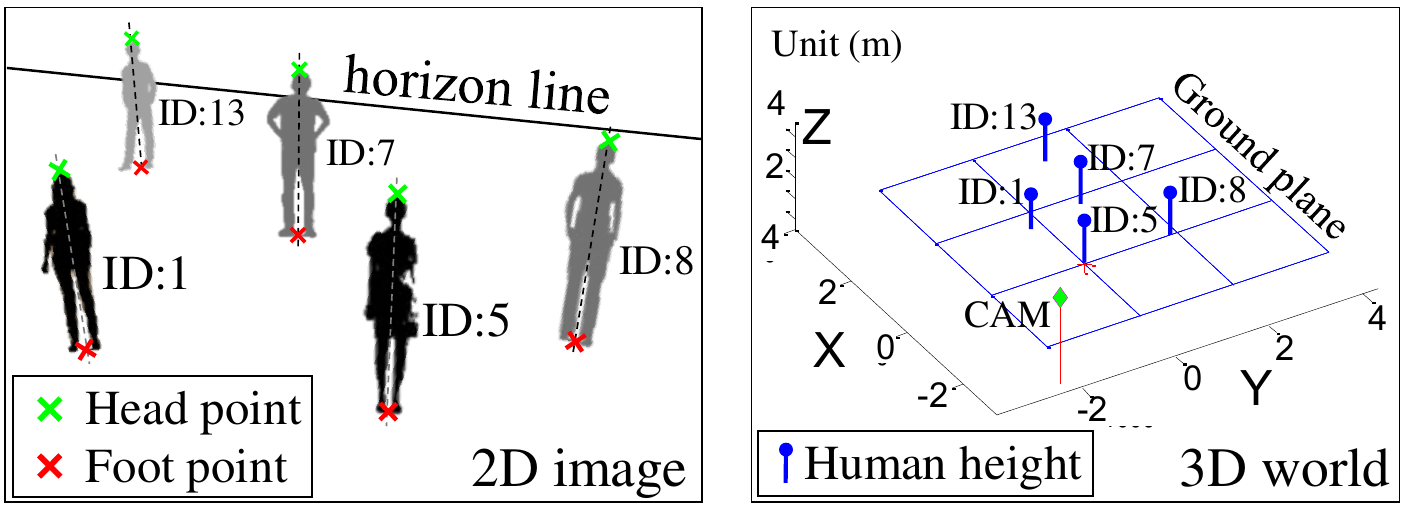}
		\caption{Examples of detected human heights in a 2D image and corresponding 3D human heights in a world coordinate system.}
		\label{fig_3}
			\vspace{-10pt}
	\end{figure}

	To this end, we estimate relative scale of human heights between cameras based on person re-identification results and adjust each camera's extrinsic parameter $\mathbf{t}$.
	We denote a person $i$ in camera $k$ as $\mathbf{o}^k_i$. Then, a matrix of people correspondences is defined as
	\begin{equation}
	\begin{split}
	\mathbf{M}^{kl} & = \left\{ \left(m_{ij}\right) |  1 \le i \le N^k ,  1 \le j \le N^l  \right\}, \\
	m_{ij} & = \begin{cases} 1\quad \text{if}~ \mathbf{o}^k_i ~\text{corresponds to~}\mathbf{o}^l_j \\ 0\quad \text{otherwise} \end{cases},
	\end{split}
	\end{equation}
	where $N^k$ and $N^l$ are the numbers of identities in camera $k$ and $l$.
	To find the relative scale between two cameras, we find a scale ratio $S$ that minimizes the following equation
	\begin{equation}
	\label{eq_3}
	S^{kl} = \underset {  S  }{ \text{arg min} }\sum _{ i=1 }^{ N^k }{ \sum _{ j=1 }^{ N^l }{ \left(  m_{ij} \left| S - \frac { { H(\mathbf{o}^k_i) } }{ { H(\mathbf{o}^l_j) } }  \right| \right) }  }, 
	\end{equation}
	where $H(\mathbf{o}^k_i)$ is an average height of a person $\mathbf{o}^k_i$ along its moving path in the world coordinate system.
	Since the proposed Eq.~\eqref{eq_3} utilizes multiple correspondence pairs, we can prevent overfitting the value $S$.
	We assume that every person is on the planar ground plane. Inspired by \cite{lv2006camera}, we can detect human foot and head points in a 2D image and compute the corresponding person's height according to the projection matrix $\mathbf{P}$ as shown in Fig.~\ref{fig_3}.
	
	After estimating the scale ratio $S$ between two cameras, the projection matrices of camera $k$ and $l$ are updated as	
	\begin{equation}
	\begin{split}
	\mathbf{P}^k & = \mathbf{K}^k\left[ \mathbf{R}^k\quad \mathbf{t}^k \right], \\
	\mathbf{P}^l & = \mathbf{K}^l\left[ \mathbf{R}^l\quad S^{kl}\mathbf{t}^l \right].
	\end{split}
	\end{equation}
	Through the proposed process, we can adjust all extrinsic camera parameters $\mathbf{t}$ in the camera network.
	Note that the proposed framework does not aim to find the absolute scale of the world coordinate system and real walking speeds of people, but to match the scales across cameras.
	
	\subsection{Distance-based Camera Network Topology Inference}
	\label{subsec:dist_top}
	
	To infer the distance-based camera network topology, we exploit the speeds of people in each camera.  
	We assume that every person is on the planar ground plane ($Z=0$, world $XY$ plane).
	Based on this assumption and camera calibration result in Sec~\ref{subsec:scale}, the speed of a person $\mathbf{o}^{k}_{i}$ is calculated as follows,
	\begin{equation}	
	\bar{ v }^{ k }_{  i   }=\frac { 1 }{ T } \sum _{ t=1 }^{ T }{ \sqrt { { \left( { X }^{ k }_{  i, t }-{ X }^{ k }_{ i, t-1 } \right)  }^{ 2 }+{ \left( { Y }^{ k }_{ i, t }-{ Y }^{ k }_{ i, t-1 } \right)  }^{ 2 } }  } ,
	\end{equation}
	where the $t$ is the time (second), and $X_{i}$,$Y_{i}$ are the world coordinates (meter) of a person $i$'s foot position in camera $k$, respectively.
	In order to get the more reliable speed of a person, we average multiple speeds of the person along its trajectory in each camera.

	\begin{figure}[t]	
		\centering
		\includegraphics[width=1\columnwidth]{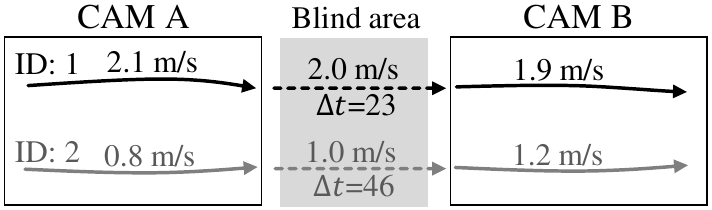}
		\caption{Example of estimating a distance between two cameras. The speed in the blind area is inferred by averaging two speeds from two cameras. The distance between cameras is estimated as $46m$ from both identities.}
		\label{fig_4}
			\vspace{-5pt}
	\end{figure}
	
	After calculating the speeds of people, we build a distance distribution based on person re-identification results between two cameras.
	For example, we have a correspondence between two cameras $\{\mathbf{o}^{k}_{i}, \mathbf{o}^{l}_{j} \}$. We then estimate the distance between two cameras by multiplying the speed and the time difference as follows,
	\begin{equation}
	\label{eq_6}
	d^{kl} = \frac{1}{2} (\bar{ v }^{ k }_{i} + \bar{ v }^{ l }_{  j }) \cdot \Delta t,
	\end{equation}
	where $\Delta t$ is a transition time of the person who appears in two cameras at different times.
	Note that it is impossible to directly observe the speed of the person in the blind area (i.e., area of between cameras). For that reason, to infer the speed in the blind area, we average the two speeds ($\bar{ v }^{ k }_{i}$ and $\bar{ v }^{ l }_{  j }$) from two cameras.
	Figure~\ref{fig_4} shows the example of the distance estimation between cameras. Although the speeds of two identities are different, the estimated distances are the same.
	Using multiple correspondences between cameras, we make a histogram of the distance and normalize the histogram by dividing with the number of the correspondences. 
	We denote the distribution according to the distance between two cameras $k$ and $l$ as $p^{kl}\left(d\right)$.
	
\begin{figure}[t]	
		\centering
		\subfigure[Transition time distribution]{\includegraphics[width=0.495\columnwidth]{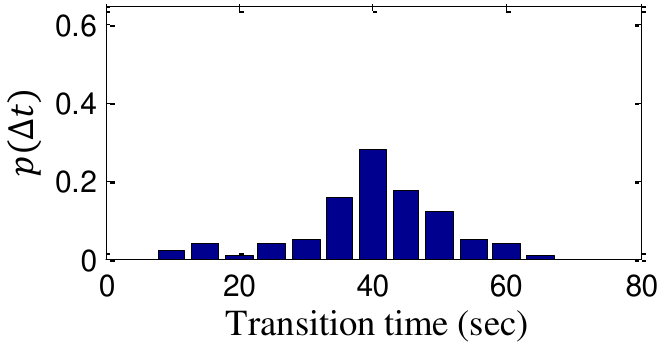}}
		\subfigure[Distance distribution]{\includegraphics[width=0.495\columnwidth]{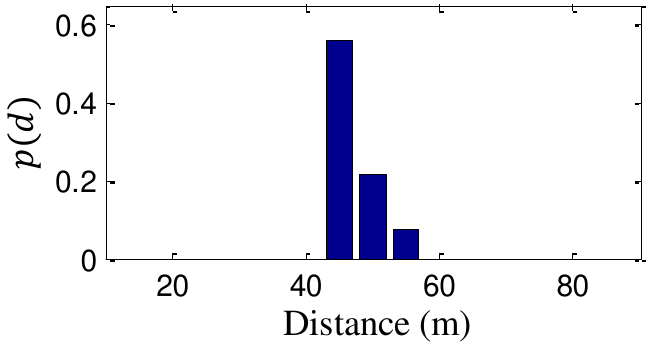}}
		\caption{Comparison of two distributions between cameras. The transition time distribution has a larger variance than that of the distance distribution.}
		\label{fig_5}
		
	\end{figure}

	By performing distance distribution estimation between all camera pairs in the camera network, we obtain the distance-based camera network topology, and it is defined by a graph as follows,
	\begin{equation}
	\begin{split}
	G_d & =\left(V, E_d\right), \\
	V & \in \left\{k|{ 1\le k \le N_{cam}  } \right\} , \\
	E_d & \in \left\{{p^{kl}\left(d\right)}|{ 1\le k\le N_{cam}, 1\le l\le N_{cam}} \right\}, 
	\end{split}
	\end{equation}
	where $N_{cam}$ is the number of cameras in the camera network and $V$ is a set of cameras and $E_{d}$ is a set of distance distribution between cameras.
	As shown in Fig.~\ref{fig_5}, the variance of the distance distribution inferred by the proposed method (Fig.~\ref{fig_5} (b)) is smaller than that of the conventional time-based transition distribution (Fig.~\ref{fig_5} (a)). Note that it is difficult to reduce the search range when the variance of the distribution is large.

	\subsection{Person Re-identification via Distance-based Camera Network Topology}
	\label{subsec:re_id}
	
	In this section, we restrict the search range based on the inferred distance-based camera network topology and perform person re-identification.
	\vspace{3pt}
	
\begin{figure}[t]	
		\centering
		\subfigure[Person speed: 2.1 m/s]{\includegraphics[width=0.495\columnwidth]{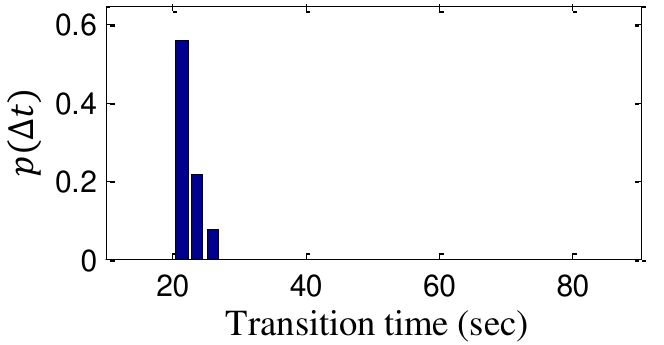}}
		\subfigure[Person speed: 0.8 m/s]{\includegraphics[width=0.495\columnwidth]{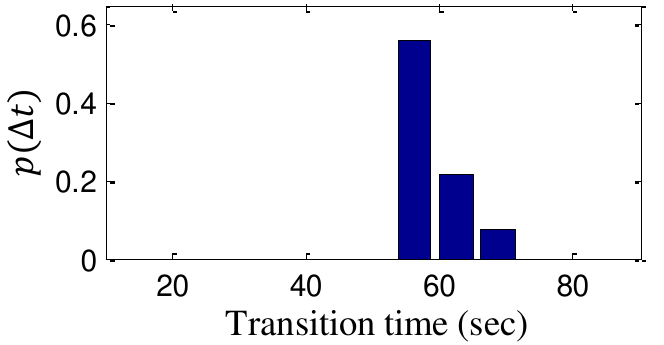}}
		\caption{Inferring adaptive transition time distribution for each identity based on the speed of a person.}
		\label{fig_6}
	\end{figure}

	\noindent \textbf{Search range restriction.} \quad	Dividing the distance distribution~$p^{kl}\left(d\right)$ by the speed of a person $i$ in camera $k$ gives a transition time distribution of the person $i$ who moves from camera $k$ to $l$ as
	\begin{equation}
	p^{kl}_{i}\left(\Delta t\right) = \frac{p^{kl}\left(d \right)}{\bar{ v }^k_{i}}. 
	\end{equation}
	Thus, we can adaptively give a camera network topology for each person depending on its speed as shown in Fig.~\ref{fig_6}.
	Based on the obtained transition time distribution $p^{kl}_{i}\left(\Delta t\right)$, we restrict the search range for re-identification as follows: \vspace{-3pt}
	\begin{itemize}
		\item Find a mean value of the transition time distribution: $m = mean\left(p^{kl}_{i}\left(\Delta t\right) \right)$.  \vspace{-3pt}
		\item Set a search range $T_r$ around the mean value to cover 95\% of the distribution: $[m-\frac{T_r}{2}, m+ \frac{T_r}{2}]$. \vspace{-3pt}
	\end{itemize}	
	In consequence, a person who moves fast (2.1 m/s) in the camera $k$ will be searched within the time range [21.4, 28.1] (sec) in the camera $l$ (Fig.~\ref{fig_6} (a)). On the other hand, a person who moves slowly (0.8 m/s) in the camera $k$ will be searched within the time range [56.3, 73.8] (sec) in the camera $l$ (Fig.~\ref{fig_6} (b)). The search range of each person is determined adaptively in this manner. 
	Thus, the proposed adaptive search strategy according to the walking speed of a person becomes more effective under the large variation of people's speed.
	
	On the other hand, a search strategy based on the conventional time-based camera network topology searches within the fixed time range [20, 60] (sec) for all test queries as shown in Fig.~\ref{fig_5} (a). Although it has the much wider time range (40 seconds) than our method, it may fail to find a correct person who moves very slowly or fast.
	For example, its search range [20, 56.3) (sec) is redundant for the person who moves slowly (0.8 m/s), and it fails to search the person if the person reappears during (60, 73.8] (sec) in the other camera.

	\vspace{3pt}
	
	\noindent \textbf{Person re-identifiction.} 
	In this work, we utilize the LOMO feature extraction method~\cite{liao2015person}, which shows promising re-identification performance, to describe the appearances of people.
	It divides a person image patch into six horizontal stripes and extracts a HSV color histogram from each stripe. It builds a descriptor based on Scale Invariant Local Ternary Pattern (SILTP). The descriptor from the 128$\times$48 (pixel) image has 26,960 dimensions.
	
	In general, each person gives multiple appearances along with its trajectory. The multiple appearances provide rich information for re-identification. However, a lot of computations are needed to take into account all the multiple appearances.
	Inspired by~\cite{zheng2015scalable}, we employ an average feature pooling method. In average pooling, the feature vectors of multiple appearances are pooled into one by the averaged summation. Therefore, we can simply compare two identities and the similarity score between two identities is defined by 
	\begin{equation}
	S\left(\mathbf{o}^k_i,\mathbf{o}^l_j\right) = {e}^{-\left\|  \Phi \left(\mathbf{o}^k_i\right) - \Phi \left(\mathbf{o}^l_j\right) \right\|_2},
	\end{equation}		
	where $\Phi\left(\cdot \right)$ is a pooled feature vector of a person. The similarity score lies on [0, 1].
	Note that it is possible to utilize any kind of feature extraction and pooling methods in our framework.	
	
	\section{Dataset and Methodology}
	\label{sec:data_meth}
	
	\subsection{Dataset}
	
	Over the past few years, numerous datasets of person re-identification have been published such as \texttt{VIPeR} \cite{gray2007evaluating}, \texttt{PRID 2011}~\cite{hirzer11a}, \texttt{CUHK}~\cite{li2013locally,li2014deepreid}, \texttt{iLIDS-VID}~\cite{wang2014person}, \texttt{MARS}~\cite{zheng2016mars} and \texttt{Airport}~\cite{Karanam2016peid_review}. 
	However, most of the datasets do not provide camera synchronization information or time stamps of all frames; thus these datasets cannot be used for testing our framework.
	
\begin{figure}[t]	
		\centering
		\includegraphics[width=0.77\columnwidth]{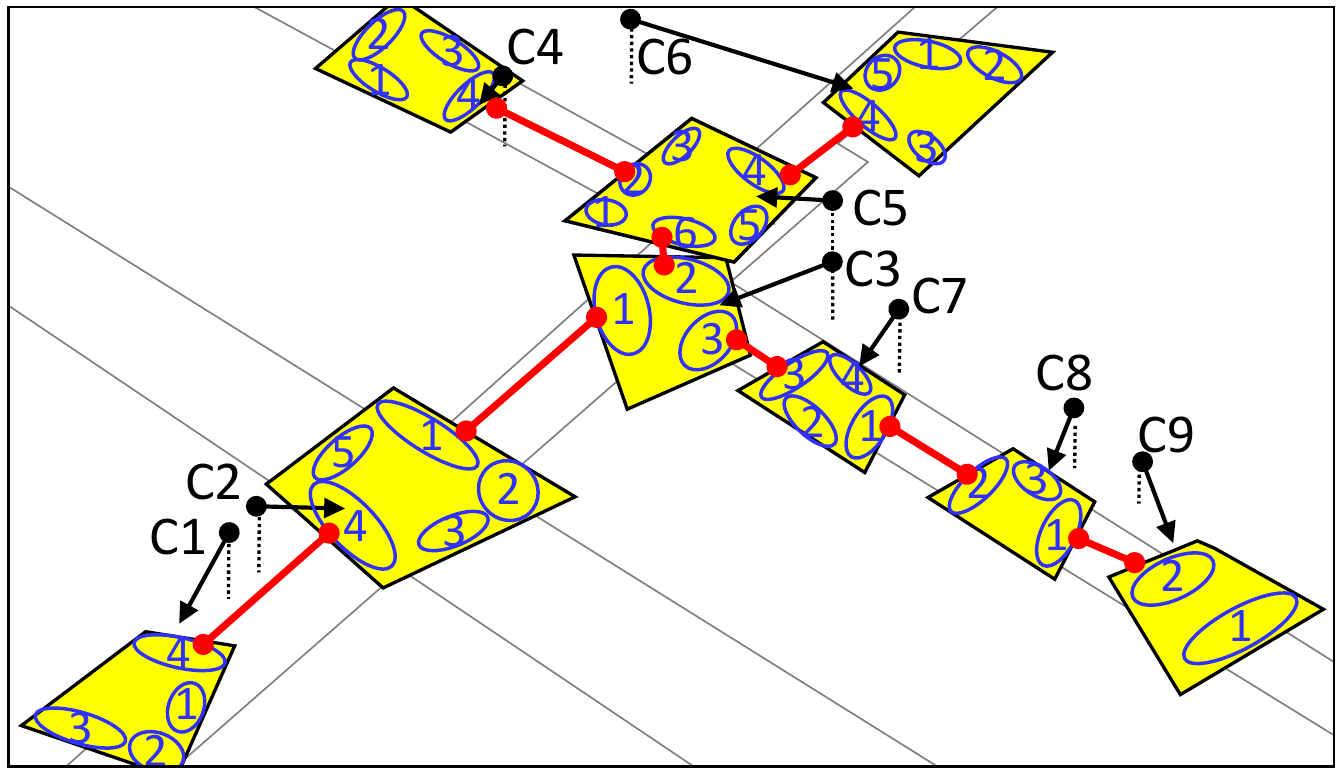}
		\caption{Layout of the \texttt{SLP} dataset. Each blue ellipse in a camera is an entry--exit zone. Red lines are valid links between cameras (best viewed in color).}
		\label{fig_7}
	\end{figure}

	To validate our methods and compare to other state-of-the-art methods, we used the \texttt{SLP} re-identification dataset~\cite{Cho_2017_ICCV_Workshops}. It is a large-scale person re-identification dataset containing nine synchronized outdoor cameras. The total number of identities in the dataset is 2,632. The layout of the camera network is shown in Fig.~\ref{fig_7}. It has eight valid links between cameras as summarized in Table.~\ref{tab_1}.
	It provides the ground truth detection and tracking information of every person including people positions (x,y locations) and sizes (height, width).
	We used the given detection and tracking results for our experiments, since we mainly focus on person re-identification and camera network topology inference problems.
	Most person re-identification researches follow this assumption and setting.
	We resized the every person image to 128$\times$48 pixels and extracted feature descriptors from the resized images.

	\begin{table}
		\centering
		\caption{Numbers of transition identities (IDs) between two cameras.}
		\label{tab_1}
		{\small
			\setlength\tabcolsep{3.5pt}  
			\begin{tabular}{c|c|c||c|c|c}
				\noalign{\hrule height 1pt}
				Link & CAM pairs  		  & \# of IDs & Link  & CAM pairs  		   & \# of IDs \\ \hline
				1    & \texttt{CAM1-CAM2} & 227        & 5    & \texttt{CAM4-CAM5} & 155       \\ \hline
				2    & \texttt{CAM2-CAM3} & 571        & 6    & \texttt{CAM5-CAM6} & 61        \\ \hline
				3    & \texttt{CAM3-CAM5} & 568        & 7    & \texttt{CAM7-CAM8} & 281       \\ \hline
				4    & \texttt{CAM3-CAM7} & 168        & 8    & \texttt{CAM8-CAM9} & 633       \\ \noalign{\hrule height 1pt}
			\end{tabular}}
		\end{table}

		\subsection{Evaluation methodology}
		
		To evaluate a person re-identification performance, many previous works plot a Cumulative Match Curve (CMC)~\cite{gray2007evaluating} representing true match being found within the first $n$ ranks. In general, to plot the CMC, the number of a gallery (i.e., matching candidates) should be fixed for all test queries. However, in our framework, the number of a gallery varies according to the camera network topology and test queries. Therefore, we cannot plot a complete CMC.
		In this work, we followed Cho's~\cite{Cho_2017_ICCV_Workshops} re-identification evaluation metric: measuring the rank-1 accuracy by $100\cdot\frac{TP}{T_{gt}}$, where $TP$ is the number of true matching results and $T_{gt}$ is the total number of ground truth pairs in the camera network.
		In the \texttt{SLP} dataset, $T_{gt}$= 2,664 as summarized in Table.~\ref{tab_1}.
		In practice, rank-1 accuracy is the most important one among all ranks since other ranks ($2,3,...,n$) failed to find the correct correspondences at least one time.
		
		To evaluate the accuracy of the camera network topology inference, we draw a curve of the retrieval rate.
		The retrieval rate represents the retrieval accuracy of matching candidates derived from the camera network topology.
		For example, if the matching candidates include the true correspondence of a test query, it counts as a success, otherwise a fail.
		Naturally, when the topology gives a wide search range $T_r$, the retrieval rate becomes high since the matching candidates within the wide search range are likely to include a true correspondence. 
		For unbiased evaluations, we draw a curve of the retrieval rate according to the average search range as shown in Fig.~\ref{fig_9}.
		If there are multiple links in the camera network, we measure the retrieval rates for all links and average them to get the final rate.
		Since we have no ground-truth of the camera network topology, measuring the retrieval rate is reasonable.


		\section{Experimental Results}
		\label{sec:exp}
		
		In the experiments, we assume that the camera parameters are partially given -- camera intrinsic parameter $\mathbf{K}$ and camera rotation $\mathbf{R}$ are given, but the camera translation is given by $S\cdot\mathbf{t}$, where $S$ is unknown.
		Thus, the cameras in the camera network do not share the same world coordinate system, initially.
		To identify a pair of cameras is connected or not, we applied the connectivity check schemes in~\cite{makris2004bridging, Cho_2017_ICCV_Workshops} and we obtained initial person re-identification results based on \cite{Cho_2017_ICCV_Workshops} for our framework.
		
		Using the initial re-identification results, we perform two steps in our framework: 1) relative scale estimation between cameras, and 2) distance-based camera network topology inference. 
		However, the initial results of the re-identification between cameras may include several false correspondences leading an inaccurate topology inference result.
		Therefore, we utilized only reliable people correspondences that have high similarity scores: $S\left(\mathbf{o}^k_i,\mathbf{o}^l_j\right) > 0.7$ and excluded the false correspondences in both steps~1) and 2). We empirically set the threshold as $0.7$, but our framework does not highly depend on this threshold.
		As a result, we found seven links in the camera network. Unfortunately, we failed to find a link \texttt{CAM5-CAM6} since the number of people is small due to the long distance from the camera (Fig.~\ref{fig_7}). Also, \texttt{CAM6} is isolated from other cameras as shown in Table.~\ref{tab_1}.
		
		\begin{figure}[tbp]	
			\centering
			\subfigure[Transition time distribution]{\includegraphics[height=1.35\columnwidth]{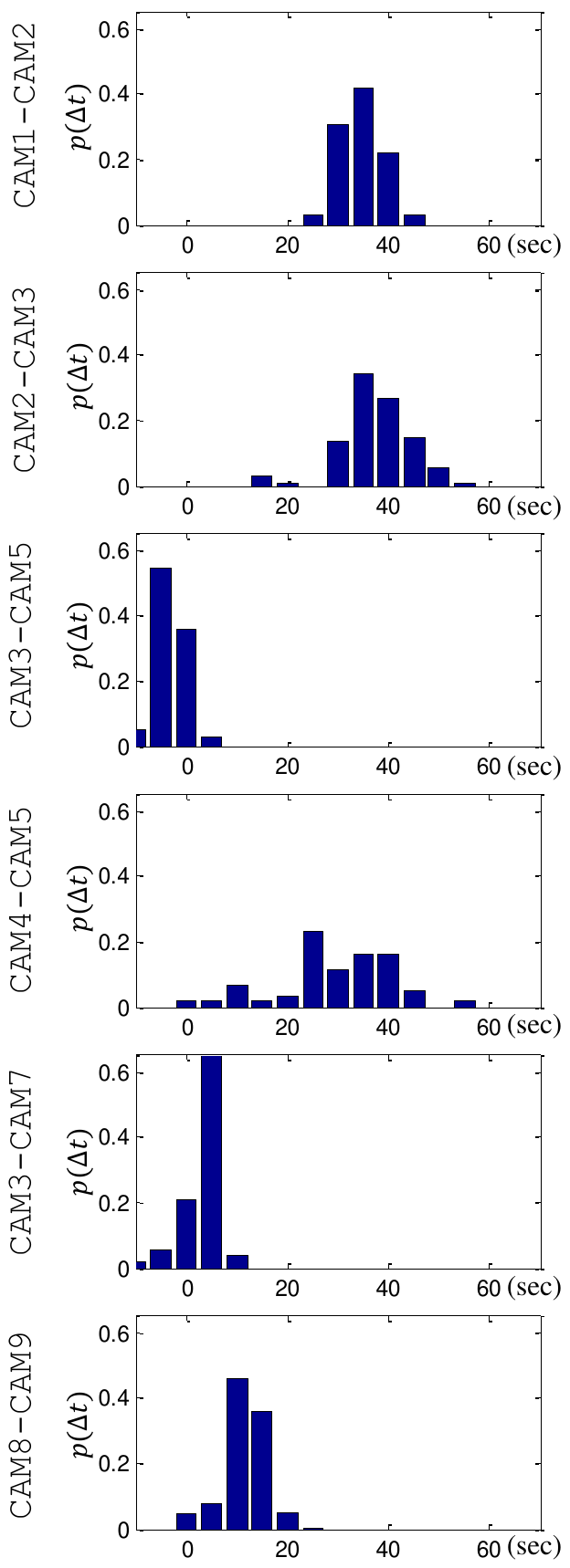}}
			\subfigure[Distance distribution]{\includegraphics[height=1.35\columnwidth]{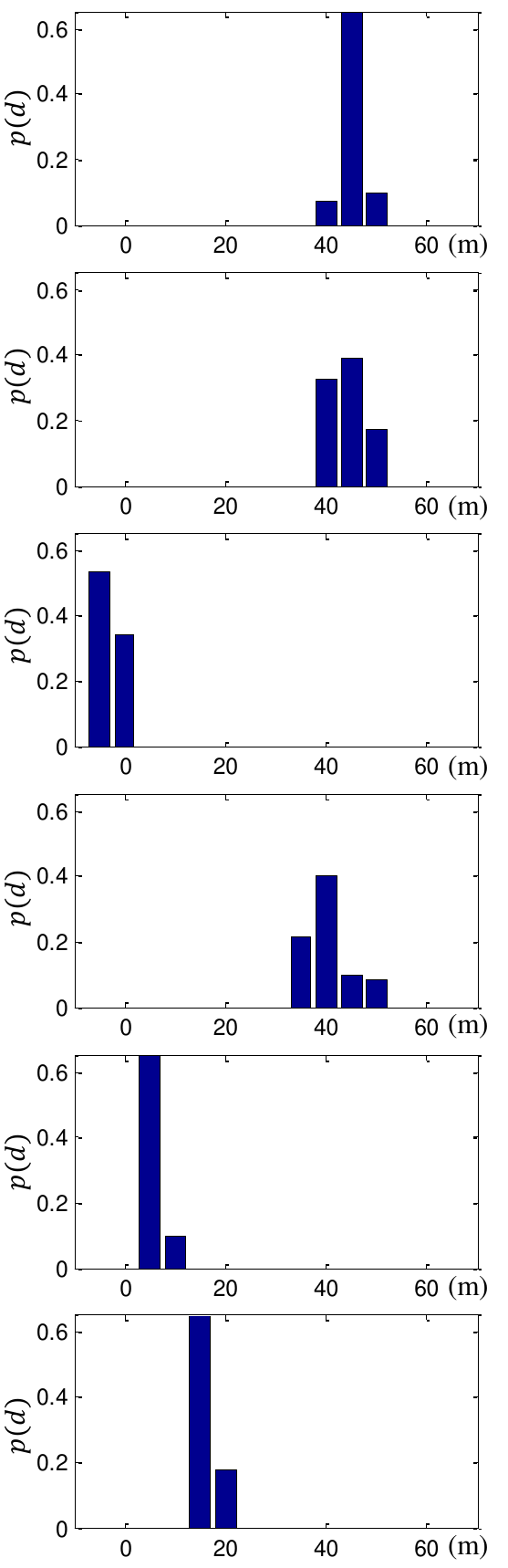}}
			\caption{Comparisons of two types of inferred distributions.}
			\label{fig_8}
				\vspace{-10pt}
		\end{figure}

		Figure~\ref{fig_8} shows comparisons of two types of distribution between cameras: (left) transition time distributions in conventional time-based topology, (right) distance distributions in proposed distance-based topology.
		As we can see, the proposed distance distributions show more clear peaks and small variances compared to those of transition time distributions.
		The result implies that the proposed distance-based topology is more effective than the time-based topology for person re-identification, since it is ambiguous to restrict search range with the unclear camera network topology.
		A \texttt{CAM3-CAM5} pair has the negative values of both transition time and distance since they are overlapped.
		
		Ideally, a distance between two cameras should be one value if there is a single path between cameras.
		However, the proposed distance distributions did not converge to one value, but have some ranges e.g., \texttt{CAM1-CAM2}: [40, 55] and \texttt{CAM4-CAM5}: [35, 55].
		This is because the speeds and paths of moving people in the blind region are totally unknown and can differ person to person.	
		In addition, other values, which are used for inferring the topology, such as camera parameters and observed people positions in each view are not perfect due to noise.
		To overcome these limitations, we estimated the distance of the blind area by interpolating the information on both sides of cameras as in Eq.~\eqref{eq_6} and could obtain quite clear distance distributions.
		
		\begin{figure}[tbp]	
			\centering
			\includegraphics[width=0.75\columnwidth]{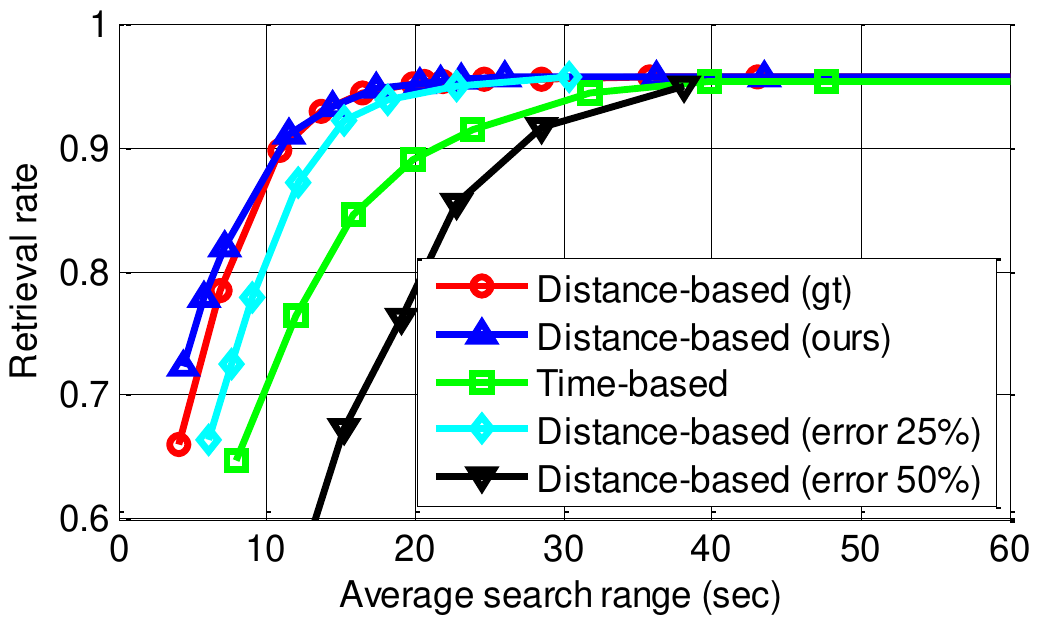}		
			\caption{Topology accuracy: retrieval rate according to the average search range.}
			\label{fig_9}
		\end{figure}	
		
		\begin{figure}[tbp]	
			\centering
			\includegraphics[width=0.75\columnwidth]{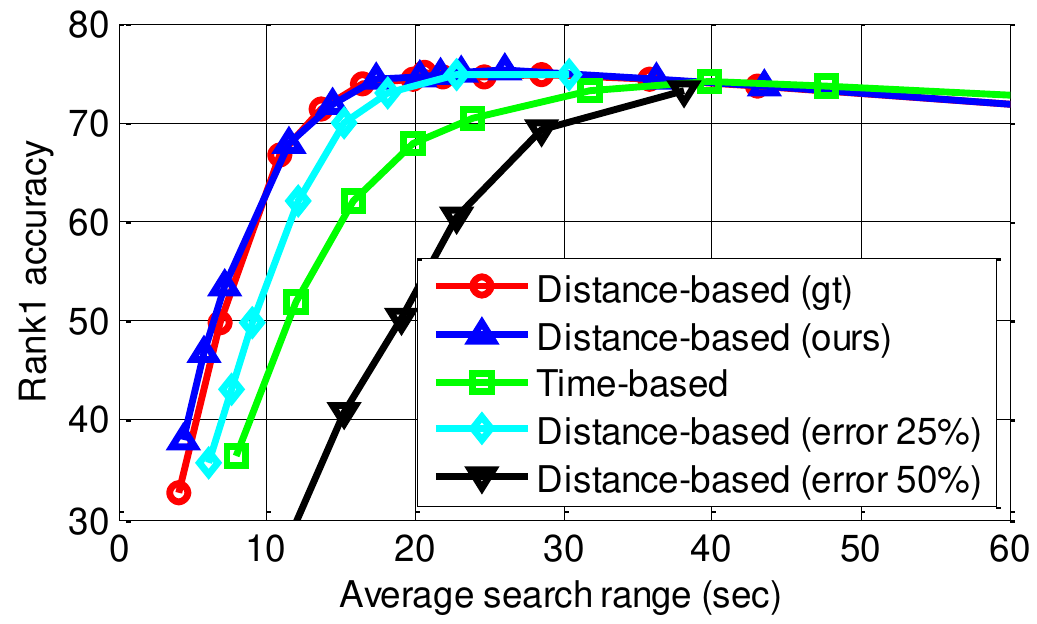}
			\caption{Re-identification accuracy: rank-1 accuracy according to the average search range.}
			\label{fig_10}	
		\end{figure}

		\begin{table*}[tbp]
			\centering
			\caption{Performance comparison with state-of-the-art methods.}
			\label{tab_2}
			\setlength\tabcolsep{4.0pt}  
			\begin{tabular}{r|c|c|c|c|c|c|c|c}
				\noalign{\hrule height 1pt}	
				Methods   & Makris's~\cite{makris2004bridging} & Nui's~\cite{niu2006recovering} & Chen's~\cite{chen2014object} & DNPR~\cite{martinel2016person} & Cai's~\cite{cai2010recovering}   & Cho's~\cite{Cho_2017_ICCV_Workshops} & ours--time & ours--dist \\ \hline 
				rank-1 accuracy   & 54.0     & 54.6    & 55.2   & 44.7   & 51.4     & 68.3     & 67.8 &  \textbf{74.7} \\ \noalign{\hrule height 1pt}	
			\end{tabular}
		\end{table*}

		\begin{figure*}[tbp]
			\centering
			\subfigure[Makris's~\cite{makris2004bridging}]                {\includegraphics[height=0.67\columnwidth]{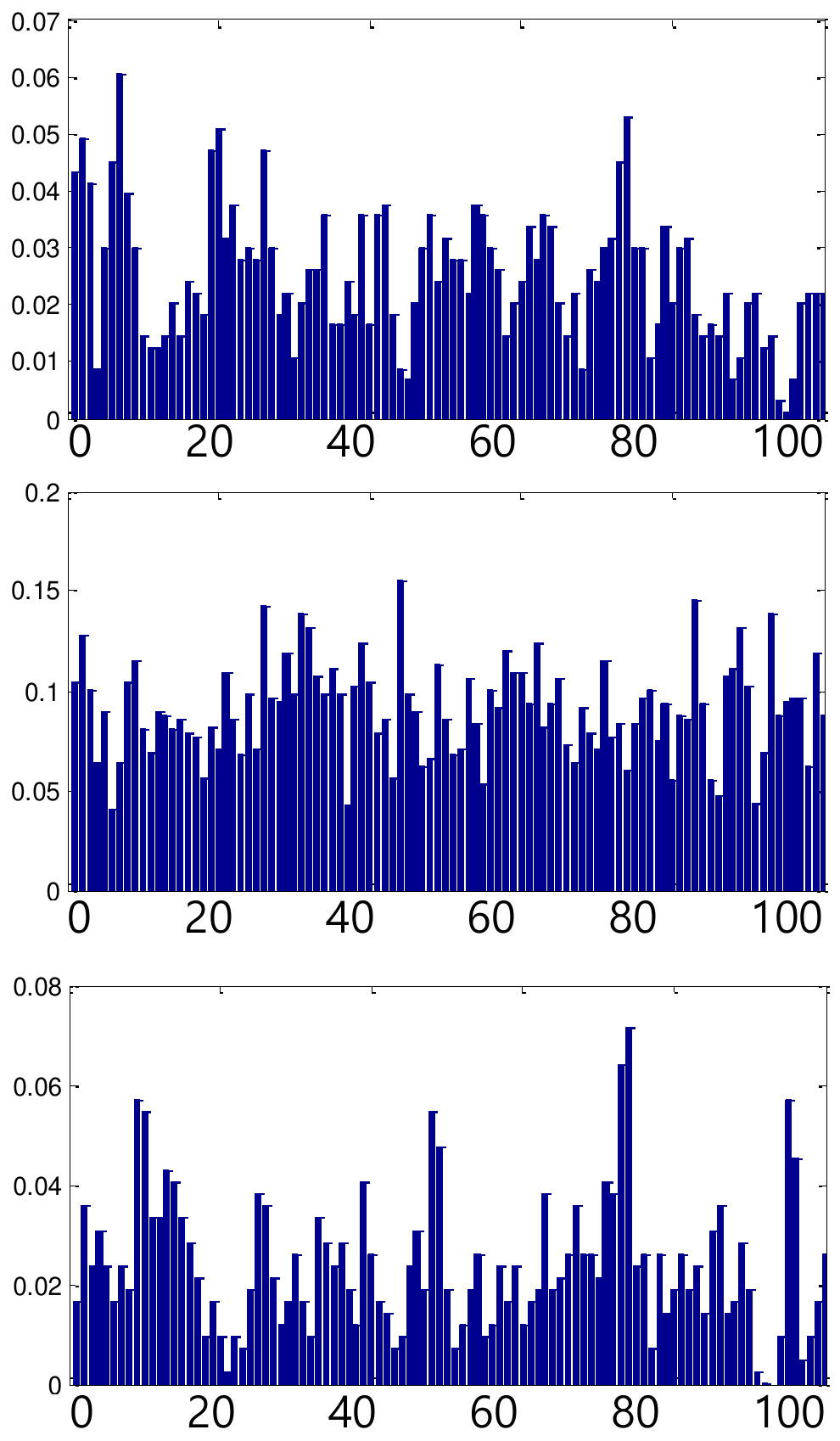}} 
			\subfigure[Nui's~\cite{niu2006recovering}]                    {\includegraphics[height=0.67\columnwidth]{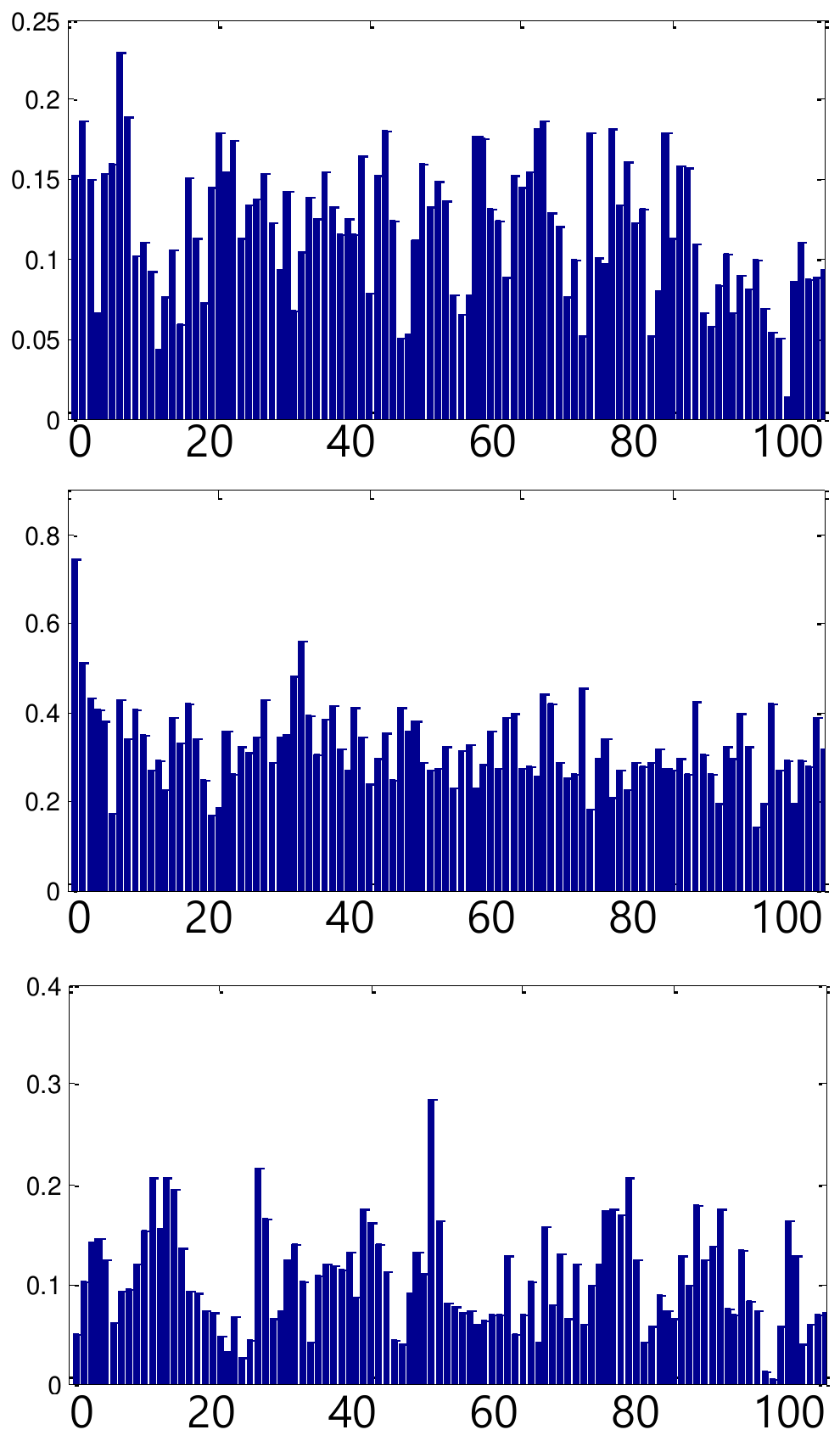}} 
			\subfigure[Chen's~\cite{chen2014object}]                      {\includegraphics[height=0.67\columnwidth]{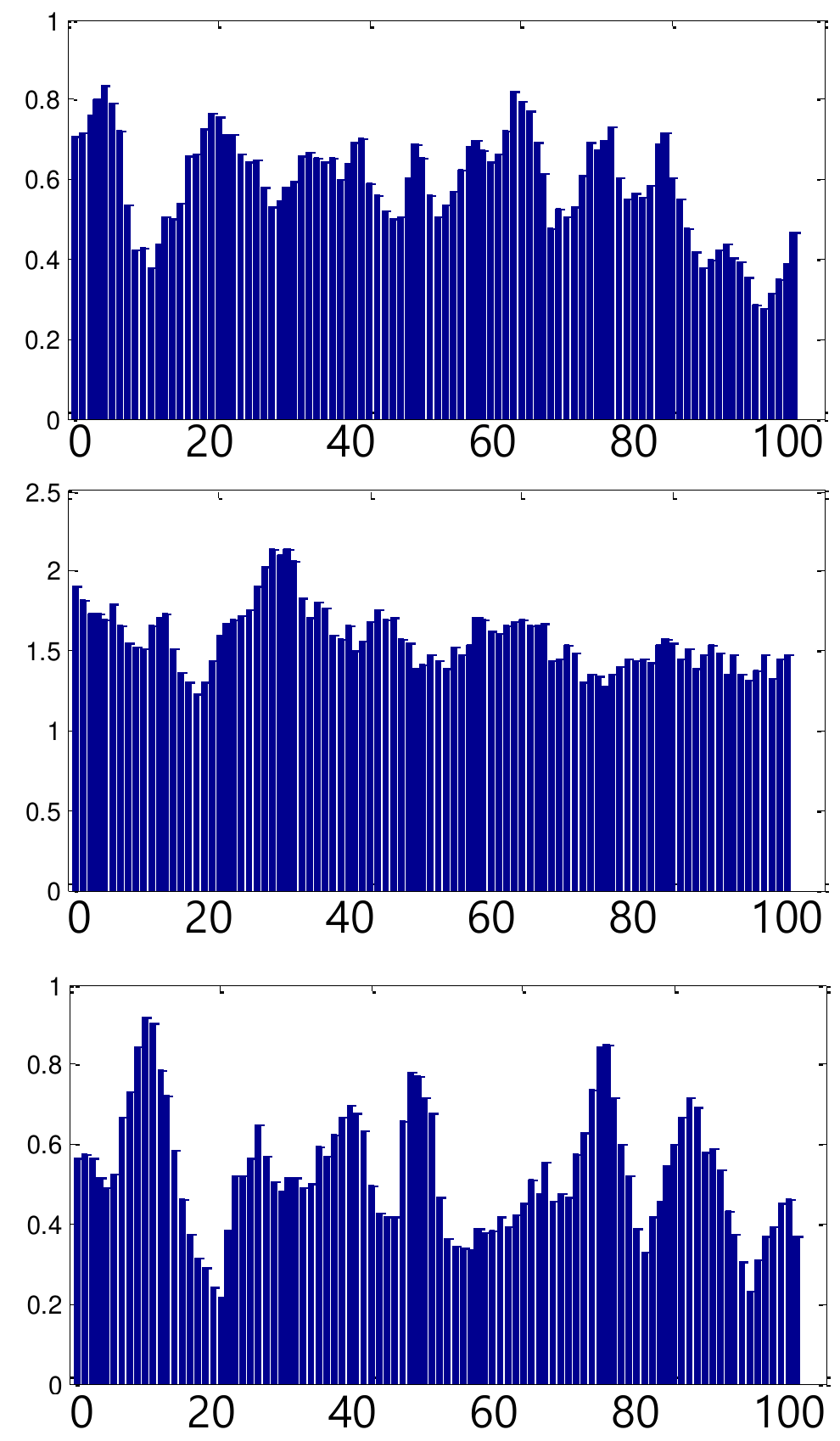}} 
			\subfigure[Cho's~\cite{Cho_2017_ICCV_Workshops}]{\includegraphics[height=0.67\columnwidth]{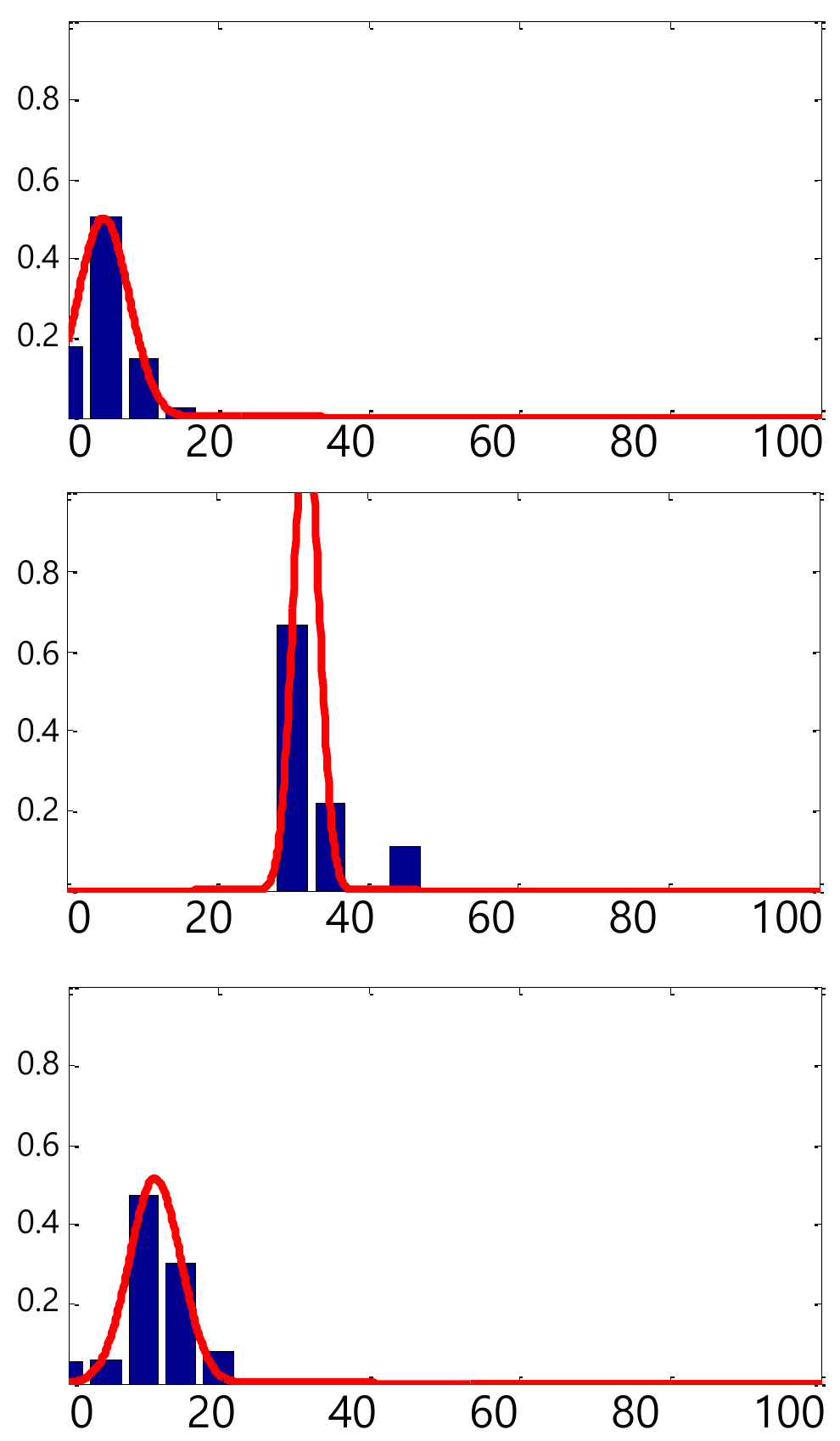}} 
			\subfigure[ours--dist]                                              {\includegraphics[height=0.67\columnwidth]{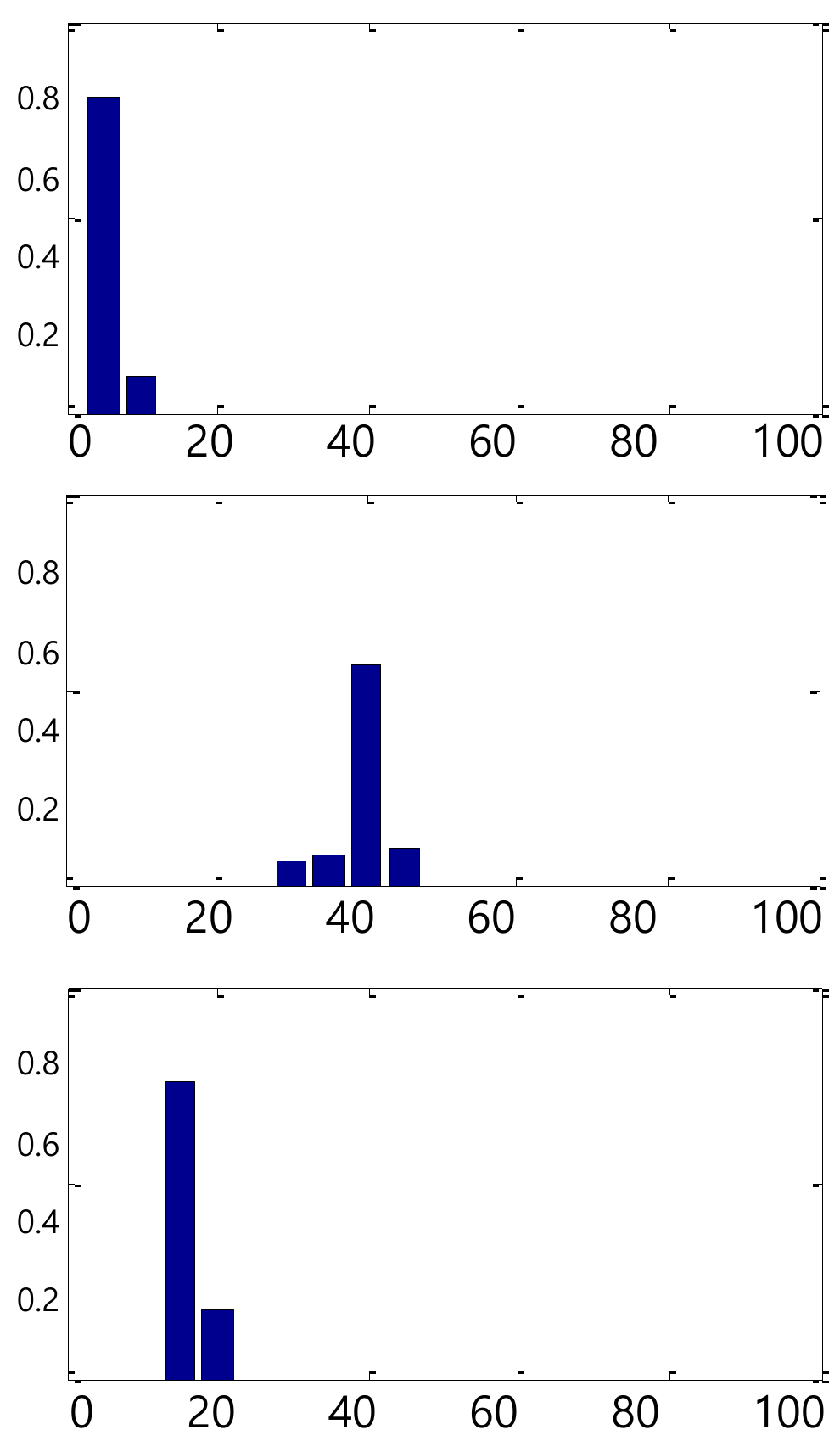}}  
			
			\caption{Comparison of inferred transition time distributions of previous methods and our distance distribution. First row -- camera pair:~\texttt{CAM3-CAM7}. Second row -- camera pair:~\texttt{CAM7-CAM8}. Third row -- camera pair:~\texttt{CAM8-CAM9}.}
			\label{fig_11}
				\vspace{-5pt}
		\end{figure*}   
		
		To validate the effectiveness of the proposed method, we first evaluated the topology inference accuracy: retrieval rate according to the average search range.
		In this experiment, we compared a distance-based method with a time-based method. For a fair comparison, we applied the same baseline~(e.g., feature extraction and pooling methods) to each method.
		As shown in Fig.~\ref{fig_9}, the proposed distance-based method (ours) shows superior performance than the conventional time-based method.
		To verify the proposed scale estimation method in Sec.~\ref{subsec:scale}, we also compared several distance-based methods with various experimental settings: $\bullet$ gt: a method using ground-truth camera parameters for all cameras, $\bullet$ error N\%: a method using camera parameters with N percent of scale error.
		As we can see, distance-based methods with erroneous camera parameters (error 25\%, 50\%) show lower performance than the distance-based method with ground-truth camera parameters (gt).	
		On the other hand, the proposed method (ours) shows a similar performance with the distance-based (gt).
		Interestingly, our method shows higher performance than distance-based (gt) around 5--10 second search times.
		In addition, a distance-based method (error 25\%) shows better a retrieval rate than that of the time-based method.
		It implies that our method is robust to camera calibration error.

		We also evaluated the accuracy of re-identification based on each camera network topology.
		As shown in Fig.~\ref{fig_10}, the proposed distance-based method (ours) shows superior re-identification performance than that of the time-based method. 
		All tested re-identification performances increase up to a certain search range and then decrease. This is because a wide search range retrieves a lot of matching candidates that are likely to include a true correspondence, but they also include lots of irrelevant identities. 
		Thus, it is important to set a proper search range.
		In our methods, we searches in the range of 20 seconds on average based on the propose search range restriction in Sec.~\ref{subsec:re_id}.
		It is reasonable for both topology and re-identification performances as shown in Fig.~\ref{fig_9} and \ref{fig_10}.

		We compared the proposed method with several previous methods~\cite{makris2004bridging, niu2006recovering, chen2014object, martinel2016person,	cai2010recovering ,	Cho_2017_ICCV_Workshops} that infer a time-based camera network topology.
		Figure~\ref{fig_11} shows comparison of inferred transition time distributions and an inferred our distance distribution. 
		As we can see, the methods~\cite{makris2004bridging, niu2006recovering, chen2014object} show very unclear and noisy transition time distributions. This is because, to infer the topology, they used a simple correlation of people exiting--entering patterns instead of utilizing re-identification results.
		On the other hand, the method~\cite{Cho_2017_ICCV_Workshops}, which used re-identification results to infer the topology, shows reasonable results than other methods~\cite{makris2004bridging, niu2006recovering, chen2014object}.
		Compared to~\cite{Cho_2017_ICCV_Workshops}, our distance distributions between camera pairs show more clear peaks and have small variances.
		
		In Table.~\ref{tab_2}, we summarize the person re-identification results based on each inferred camera network topology.
		In this experiment, we have two results: ours--time and ours--dist. They share the same baseline except for the utilized camera network topology (time-, and distance-based) for re-identification.
		For a fair comparison, we set the same average search range (20 seconds) for both of our methods.
		Among the methods, which utilized the time-based camera network topology, the method~\cite{Cho_2017_ICCV_Workshops} showed the highest re-identification performance.
		It employed the random forest algorithm~\cite{breiman2001random} to perform accurate person re-identification.
		On the other hand, our methods (ours--time, ours--dist) used a simple feature pooling method for re-identification as described in Sec.~\ref{subsec:re_id}.
		Although our method employed the simpler re-identification method, re-identification using the proposed distance-based topology shows superior performance than other state-of-the-art methods.

		\section{Conclusions}
		\label{sec:conclusion} 
		
		In this paper, we proposed a novel distance-based camera network topology inference. 
		We first estimate relative scale ratio between cameras based on the human heights information and infer the distance-based camera network topology. The proposed distance-based topology can be applied adaptively to each person according to its speed; therefore it can effectively handle the various people transition time between cameras. In order to validate the proposed method, we used a public synchronized large-scale re-identification dataset and compared our method with state-of-the-art methods. 
		The results show that the proposed method is promising for person re-identification in large-scale camera network with various people transition time between cameras.
		
		{\small
			\bibliographystyle{ieee}
			\bibliography{egbib}
		}

	\end{document}